\documentclass[conference]{IEEEtran}

\usepackage{cite}
\usepackage{amsmath,amssymb,amsfonts}
\usepackage{graphicx}
\usepackage{textcomp}
\usepackage{xcolor}
\usepackage{url}

\def\BibTeX{{\rm B\kern-.05em{\sc i\kern-.025em b}\kern-.08em
    T\kern-.1667em\lower.7ex\hbox{E}\kern-.125emX}}

\usepackage{listings}
\usepackage{xspace}
\usepackage{subfig}
\usepackage{todonotes}
\usepackage{tablefootnote}
\usepackage{algorithm}
\usepackage{algpseudocode}

\usepackage{booktabs}
\usepackage{multirow}
\usepackage{threeparttable}
\usepackage{tikz}
\usepackage{amsthm}
\usepackage{environ}

\newcommand{\fix}[1]{}
\newcommand{\comm}[1]{\textcolor[rgb]{0.0,0.0,1.0}{#1}}
\newcommand{\method}{AxoNN}

\NewEnviron{NORMAL}{%
    \scalebox{1.25}{$\BODY$} 
}

\definecolor{dkgreen}{RGB}{0,64,0}
\definecolor{ltgray}{RGB}{245,245,245}
\definecolor{mauve}{RGB}{139,0,139}

\begin{document}

\title{A 4D Hybrid Algorithm to Scale Parallel Training to Thousands of GPUs}

\author{\IEEEauthorblockN{Siddharth Singh, Prajwal Singhania, Aditya K.~Ranjan, Zack Sating, Abhinav Bhatele}
\IEEEauthorblockA{~\\
    \textit{Department of Computer Science}\\
    \textit{University of Maryland}\\
    E-mail: \{ssingh37, prajwal, aranjan2, zsating\}@umd.edu, bhatele@cs.umd.edu
}
}

\maketitle

\begin{abstract}
Heavy communication, in particular, collective operations, can become a
critical performance bottleneck in scaling the training of billion-parameter
neural networks to large-scale parallel systems. This paper introduces a
four-dimensional (4D) approach to optimize communication in parallel training.
This 4D approach is a hybrid of 3D tensor and data parallelism, and is
implemented in the \method framework.  In addition, we employ two key
strategies to further minimize communication overheads. First, we aggressively
overlap expensive collective operations (reduce-scatter, all-gather, and
all-reduce) with computation. Second, we develop an analytical model to
identify high-performing configurations within the large search space defined
by our 4D algorithm. This model empowers practitioners by simplifying the
tuning process for their specific training workloads. When training an
80-billion parameter GPT on 1024 GPUs of Perlmutter, \method surpasses
Megatron-LM, a state-of-the-art framework, by a significant 26\%. Additionally,
it achieves a significantly high 57\% of the theoretical peak FLOP/s or 182
PFLOP/s in total.

\end{abstract}

\begin{IEEEkeywords}
Parallel deep learning, Tensor parallelism, Communication modeling, Asynchronous communcation
\end{IEEEkeywords}

\section{Introduction}
\label{sec:intro}
The effectiveness of deep learning (DL) models at generalization improves
reliably with an increase in their number of
parameters~\cite{belkin:double-descent, scaling-laws}. This trend has led to
the development of large foundational models trained using deep neural networks
(DNNs) with hundreds of billions of parameters~\cite{gpt-3,
megatron-turing-nlg-530b}.  Given the substantial memory requirements for
training these models, which often exceeds the memory capacity of single
server-class GPUs, the use of GPU-based clusters for model training has become
common. Consequently, it is imperative to develop efficient parallel algorithms
and frameworks that can leverage the combined memory capacity and computational
power of hundreds to thousands of GPUs for efficient training of such neural
networks.

The foremost challenge in scaling parallel DL training across multi-GPU
clusters is communication.  While modern GPUs have significantly improved
compute efficiency due to their use of specialized cores such as Tensor Cores
in NVIDIA GPUs, network bandwidth across compute nodes has lagged behind. This
results in modern frameworks for parallel DL training being inefficient at
scale due to the considerable overhead of message passing. These overheads stem
primarily from two factors: (i) the inherently large communication volume
associated with the underlying parallel DL algorithms, and (ii) inefficient use
of message passing with minimal to no overlap with computation. These
communication challenges deteriorate the efficiency of parallel frameworks,
becoming progressively more severe as we scale to thousands of GPUs.
Unfortunately, scalability is increasingly crucial due to the compute-intensive
nature of modern training workloads such as large language models (LLMs). 

To overcome challenges (i) and (ii), we propose a four-dimensional (4D) hybrid
parallel algorithm, implemented in \method, a framework for parallel DL
training. This 4D approach is a hybrid of 3D tensor and data parallelism.  We
use a variation of Agarwal et al.'s parallel matrix multiplication
algorithm~\cite{agarwal-3d} to efficiently parallelize the compute-intensive
matrix multiplications within each layer of the neural network. This is often
referred to as tensor parallelism. While utilizing an efficient parallel matrix
multiplication algorithm is a crucial step, there are other factors needed to
achieve communication efficiency. To further improve performance, we employ the
following two approaches.

\vspace{0.05in}
\noindent\emph{Overlapping Communication and Computation}: Many tensor parallel
approaches, including the one proposed in this work, rely on collective
communication operations (reduce-scatter, all-gather, and all-reduces). These
operations can be expensive at scale in terms of performance. To address
scalability issues, we propose several communication optimizations that
leverage asynchronous communication primitives, while being mathematically
equivalent to the synchronous implementation. These primitives allow for
significant overlap between communication and computation, maximizing hardware
utilization.

\vspace{0.05in}
\noindent\emph{Communication-aware Configuration Selection:} Our 4D algorithm
requires arranging the available GPUs in a virtual 4D grid and deciding the
sizes of each dimension. The distribution of data and compute work on this
virtual 4D grid can significantly impact communication costs. To assist users,
we introduce a model that identifies a small set of communication-optimal
configurations for a given DL workload, eliminating the need to explore the
entire search space of possible values for each of the 4 dimensions.  Rather,
users can profile these suggested configurations, streamlining the process of
finding the optimal settings for their specific workload.

We demonstrate the performance of our framework by conducting scaling studies
on multi-billion parameter DNNs, then comparing our performance with three
state-of-the-art parallel deep learning frameworks --
Megatron-LM~\cite{megatronlm} and DeepSpeed-3D~\cite{zero_3D}, and
ZeRO-3~\cite{sc2020zero}, on both the Perlmutter (Nvidia A100 GPUs) and
Frontier (AMD MI250X GPUs) supercomputers.  In a weak scaling study using GPT-3
transformer models ranging 5B-80B parameters over 64-1024 GPUs, we observe
significant performance improvements of 25--45\% over Megatron-LM and 32--50\%
over DeepSpeed-3D on Perlmutter, and 23--35\% over DeepSpeed-3D on Frontier. We
demonstrate that~\method scales well, even on AMD GPUs, where other frameworks
struggle. We also show significant improvements when training vision models
(UNet CNNs) in a weak scaling study when compared to ZeRO-3~\cite{sc2020zero}.

In summary, we make the following contributions:
\begin{itemize}
    \item An open-source, 4D tensor + data hybrid parallel framework, \method,
which reduces communication overhead and speeds up DNN training at scale,
compared to other state-of-the-art frameworks.
    \item Techniques for optimizing collective communication by leveraging
asynchrony and intelligent communication scheduling, which maximize overlap
between computation and communication.
    \item A performance model for communication, tailored to assist users in
discovering communication-minimizing configurations for~\method.
\end{itemize}

\section{Related Work}
\label{sec:related}
In this section, we present related work on different frameworks and algorithms
for parallel deep learning training, primarily focusing on tensor parallelism
and communication performance modeling.

\subsection{Tensor Parallelism} \label{sec:bg-tp}

Tensor parallel algorithms work by parallelizing the computation of every layer
of the neural network. Most frameworks for tensor parallelism focus on
fully-connected (FC) and/or convolutional layers.  The most widely used tensor
parallel framework is Shoeybi et al.'s Megatron-LM~\cite{megatronlm}. In their
work the authors propose an algorithm to parallelize a pair of FC layers.  They
apply their technique to parallelize large GPT style transformers efficiently
within GPUs in a node. 

Qifan et al.~propose a 2D tensor parallel algorithm for FC layers~\cite{you-2d}
based on the SUMMA algorithm for distributed matrix multiplication.  Similarly,
Wang et al.~propose a 2.5D parallel algorithm for FC layers~\cite{you-2.5d}.
Perhaps the closest algorithm to our work is Bian et al.'s 3D tensor parallel
algorithm~\cite{you-3d}, which is also based on Agarwal's 3D matrix
multiplication algorithm. However, (i) the authors do not propose any
communication overlap optimizations like we do in Section~\ref{sec:opt}, and
(ii) they do not provide any discussion on choosing the optimal 3D
configurations for their algorithm and instead heuristically opt for symmetric
cubic configurations.  As we show in Section~\ref{sec:comm-model}, arriving
upon optimal/near-optimal configurations is very critical for performance. Also
they only show results on single layers, whereas we demonstrate results on full
fledged multi-billion parameter models on 1000s of GPUs.

Jangda et al.~develop high performance GPU kernels that overlap computation
with communication in Megatron-LM's algorithm~\cite{jangda2022breaking}.
Dryden et al.~propose channel and filter parallelism for convolution
layers~\cite{channel-filter-parallel-cnn-dryden}. Wang et al.~propose using
asynchronous sends instead of all-gather operations for a 2D tensor parallel
scheme to overlap communication and computation \cite{wang2023overlap}. Merak
\cite{merak} introduces an automated 3D parallel framework based on graph
partitioning, along with techniques to overlap communication with computation
in pipeline and tensor parallelism modes. Li et al.~propose Oases, which
overlaps backward pass communication with activation
recomputation~\cite{li2023automated-oases}. 

\subsection{Modeling Communication Performance}
\label{sec:bg-commmodeling}

In order to alleviate the complexity of choosing the correct mapping of GPUs to
the different parallelism dimensions, several works have proposed automated
frameworks that try to model the behavior of the configurations with respect to
the communication and computation costs.  

Alpa~\cite{alpa} is a compiler that automates the process of parallelizing
neural networks by coming up with communication efficient strategies for
decomposing a given set of GPUs into a hybrid 1D tensor, pipeline and data
parallelism scheme.  However, (i) they only model a 1D tensor parallel
approach, whereas our communication model accounts for a 3D tensor parallel
paradigm (see Section~\ref{sec:comm-model}), and (ii) their communication model
is placement-agnostic and only models the communication volume. In contrast,
the communication model proposed in this work is placement-aware and accounts
for variations in bandwidths depending on the mapping of the process groups to
the underlying topology.

Cheng et al.~develop a hierarchical communication matrix over a 2-dimensional
device mesh to model the communication cost~\cite{cheng2023atp}, taking the
underlying network topology into account, and use it to automate the
decomposition over a 2D tensor parallelism scheme. Li et al.~extend Alpa and
model the cost of overlapped communication-computation for improving the
automated parallel plan~\cite{li2023automated-oases}. Alok et al.~propose
parallel algorithms and model communication costs for training Graph Neural
Networks~\cite{tripathy2020reducing}.

\section{Designing a Hybrid 3D Tensor and Data Parallel Framework}
\label{sec:design}
In this section, we describe our new approach to scaling the training of large
multi-billion parameter neural networks to thousands of GPUs. We have designed
a hybrid parallel approach that combines 3D tensor and data parallelism. Below,
we describe both components, starting with data parallelism.

\subsection{Data Parallelism}
\label{sec:design-data}

Let us assume that we want to parallelize training on $G$ GPUs. When using only
data parallelism, we first instantiate a full copy of the neural network on
every GPU, and then divide the input batch into equal-sized {\em shards} among
these GPUs. However, since we want to use a hybrid approach that combines data
with tensor parallelism, we first organize the total number of GPUs, $G$, into
a virtual 2D grid, $G_{\mathrm{data}} \times G_{\mathrm{tensor}}$. This results
in $G_{\mathrm{data}}$ groups of $G_{\mathrm{tensor}}$ GPUs each. We use data
parallelism across the $G_{\mathrm{data}}$ groups, and tensor parallelism
within each group. Similar to pure data parallelism, the $G_{\mathrm{data}}$
groups in hybrid parallelism also have to synchronize their weights by issuing
all-reduces on their gradients after every batch.

\subsection{Three-dimensional Tensor Parallelism}

Next, we describe how each GPU group, composed of $G_{\mathrm{tensor}}$ GPUs,
parallelizes the work within their copy of the neural network. Each GPU group
processes the batch shard assigned to them. Tensor parallelism refers to
parallelizing the computation within every layer of the neural network across
GPUs. We first describe the parallelization of a single layer using our
approach. We use the fully-connected (FC) or Linear layer as an example.

\begin{figure}[h]
    \centering
      \includegraphics[width=3in]{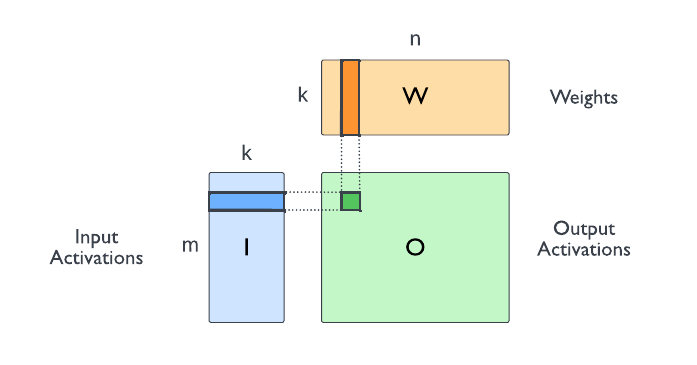}
      \caption{Computation in the forward pass of a fully-connected (FC) layer with input $I$ and layer weights $W$. 
      The output, $O$ is a matrix multiplication of $I$ and $W$. We assume $I \in \mathbb{R}^{m \times k}$, 
      $W \in \mathbb{R}^{k \times n}$, and  $O \in \mathbb{R}^{m \times n}$. \label{fig:schematic-fc}}
\end{figure}

Let us first look at the serial computation in an FC layer. Each FC layer
computes one half-precision matrix multiplication in the forward pass and two
half-precision matrix multiplications in the backward pass. The inputs to the
matrix-multiply (MM) kernel in the forward pass are the input activation, $I$,
and the layer's weight matrix, $W$. The output of the MM operation is the
output activation, $O$. This is illustrated in Figure~\ref{fig:schematic-fc}.
In the backward pass, there are two MM operations, $\frac{\partial L}{\partial
O} \times W^{\top}$ and $I^{\top} \times {\frac{\partial L}{\partial O}}$,
where $L$ is the training loss.  Thus, parallelizing an FC layer requires
parallelizing these three MM operations across multiple GPUs.

\begin{figure}[h]
    \centering
      \includegraphics[width=\columnwidth]{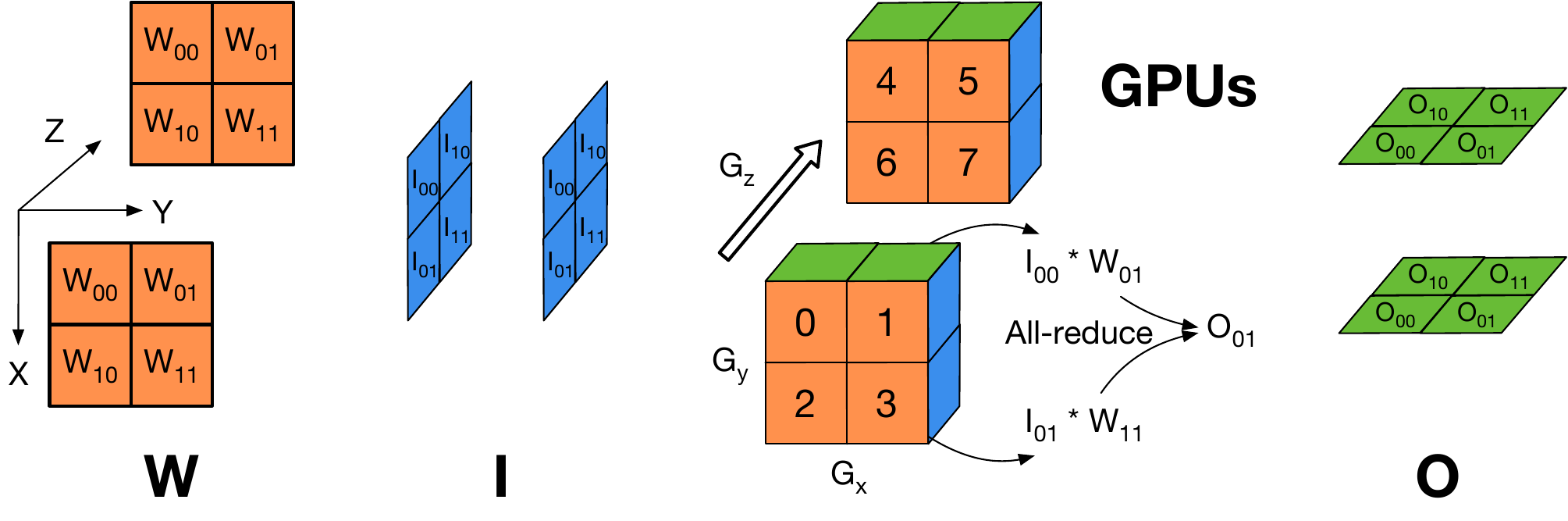}
      \caption{Parallelization of an FC layer with Agarwal's 3D parallel matrix
multiplication algorithm~\cite{agarwal-3d} on eight GPUs organized in a
$2\times2\times2$ topology. We use $G_x$, $G_y$, and $G_z$ to refer to
the number of GPUs along the three dimensions of the virtual grid topology.}
      \label{fig:schematic-agarwal-data-dist}
\end{figure}

In order to parallelize a single matrix-multiply computation across several
GPUs, we adapt Agarwal et al.'s 3D parallel matrix multiplication
algorithm~\cite{agarwal-3d}.  As noted in Section~\ref{sec:design-data}, we
need to exploit $G_{\mathrm{tensor}}$ GPUs for tensor parallelism within each
group.  Since Agarwal's algorithm uses a virtual 3D grid of processes, we first
organize the $G_{\mathrm{tensor}}$ GPUs further into a virtual
three-dimensional (3D) grid of dimensions $G_x \times G_y \times G_z$.  As an
example, we show a topology of eight GPUs with $G_x = G_y = G_z = 2$ in
Figure~\ref{fig:schematic-agarwal-data-dist}.  Additionally, we use $g_{i,j,k}$
to refer to a GPU in the grid.

Now let us discuss how we use Agarwal's algorithm to distribute input
activations, $I$, and weights, $W$, onto this 3D grid of GPUs. We do 2D
decompositions of both $I$ and $W$ into sub-blocks and map them to orthogonal
planes of the 3D grid.  For example, in
Figure~\ref{fig:schematic-agarwal-data-dist}, we observe that $W$ is
partitioned along the $X$ and $Y$-axes, and replicated along the $Z$-axis. This
means that GPUs groups in each $XY$ plane have a copy of $W$. The $I$ matrix on
the other hand is partitioned along the $X$ and $Z$-axes, and replicated along
the $Y$-axis. Once each GPU has a unique sub-block of I and W, it can compute a
portion of the $O$ matrix, which can be aggregated across GPUs in the $Y$
direction using all-reduces.

In our adapted version of Agarwal's algorithm, instead of replicating $W$ along
the $Z$-axis, we further shard $W$ along the $Z$-axis and denote these
sub-shards as $\hat{W}$. This is done to save memory as the set of GPUs along
the $Z$-axis will only have to store the gradients and optimizer states of
unique shards of the weights. We now discuss how we adapt Agarwal's algorithm
to work with sharded weight matrices in the forward and the backward passes of
our 3D tensor parallel algorithm. We illustrate the forward pass in function
\texttt{TENSOR\_PARALLEL\_FORWARD\_PASS} of Algorithm~\ref{alg:3d-tensor} from
the perspective of GPU $g_{i,j,k}$. 

\begin{algorithm}[h]
    {\small 
    \caption{Our 3D tensor parallelism for ${g}_{i,j,k}$ in a $G_{x} \times G_{y} \times G_{z}$ grid. We highlight all communication 
    operations in blue. \label{alg:3d-tensor}} 
    \begin{algorithmic}[1]
    \setlength{\lineskip}{5pt}
    \Function{tensor\_parallel\_forward\_pass}{$I_{k,j}$, $\hat{W}_{j,i}$}
        \State  $W_{j,i}$ = $\Call{\comm{${\text{all-gather}}_{z}$}}{\hat{W}_{j,i}}$
        \State  $\hat{O}_{k,i} = I_{k,j} \times {W}_{j,i}$
        \State $O_{k,i}$ $\gets$ \Call{\comm{${\text{all-reduce}}_{y}$}}{$\hat{O}_{k,i}$}
        \State // Cache $I_{k,j}$ and $W_{j,i}$  for the backward pass
        \State \Return $O_{k,i}$
    \EndFunction
    \State
    \Function{tensor\_parallel\_backward\_pass}{$\frac{\partial L}{\partial O_{k,i}}$}
        \State Retrieve  $I_{k,j}$ and $W_{j,i}$  from cache 
        \State $\frac{\partial L}{\partial I_{k,j}}$ $\gets$ \Call{\comm{${\text{all-reduce}}_{x}$}}{$\frac{\partial L}{\partial O_{k,i}} \times {W}^{\top}_{j,i}  $}
        \State ${\frac{\partial L}{\partial \hat{W}_{j,i}}}$ $\gets$ \Call{\comm{$\text{reduce-scatter}_{z}$}}{$ I^{\top}_{k,j}  
        \times {\frac{\partial L}{\partial O_{k,i}}} $}
        \State \Return $\frac{\partial L}{\partial I_{k,j}}$, ${\frac{\partial L}{\partial \hat{W}_{j,i}}}$
    \EndFunction
    \end{algorithmic}
    }
\end{algorithm}

The inputs to this function are $I_{k,j}$ and $\hat{W}_{j,i}$ i.e. the shards
of $I$ and $W$ mapped to GPU $g_{i,j,k}$ by our algorithm.  Since we have
performed an extra sharding of $W$ along the $Z$-axis, we first bring back the
full required sub-block of $W$ by issuing an all-gather on $\hat{W}$'s to get
$W$ (line 2). Then, every GPU computes a matrix multiply of their local
partitions of the input activations and weights, which is $I_{k,j} \times
W_{j,i}$ (line 3).  However, since the columns of $I$ are distributed across
the GPUs along the $Y$-axis, this step requires an all-reduce operation within
the $Y$- tensor parallel GPUs to compute the complete output (line 4).
Finally, at the end of the forward pass, each GPU caches its local partitions
of $I$ and $W$, as these are required in the backward pass (line 5).  

Let us now discuss the parallelization in the backward pass.  $\frac{\partial
L}{\partial O_{k,i}}$ is the partial derivative of the loss with respect to the
output of the forward pass, which serves as the input to the backward pass.
First, we retrieve the local partitions of the data, which we had cached
earlier in the forward pass (line 10).  After this step, we have all the data
in place to begin computing the two matrix multiplications in the backward
pass. We start with computing the gradients of the loss with respect to $I$
i.e.  $\frac{\partial L}{\partial I}=\frac{\partial L}{\partial O} \times
W^{\top}$.  For this, each GPU does a matrix multiplication, $\frac{\partial
L}{\partial O_{k,i}} \times {W}^{\top}_{j,i}  $ (line 11). Just like the
forward pass, this results in a partial output which needs to be aggregated via
an all-reduce. However, in this case the all-reduce is done by GPUs along the
$X$-axis (line 11). Next, we compute the derivative with respect to the weights
by multiplying the transpose of the local partition of $I$ with the local
partition of $\frac{\partial L}{\partial O}$ i.e. $I^{\top}_{k,j}  \times
{\frac{\partial L}{\partial O_{k,i}}} $. Finally, we do a reduce-scatter on the
outputs so that each GPU ends up with the gradients of their shard of the
weights (line 12). 

\vspace{0.07in}
\noindent\textbf{Extension to convolution layers:}
Algorithm~\ref{alg:3d-tensor} can be easily extended to convolution layers by
treating $k$ and $n$ as the number of input and output channels, respectively.  

\vspace{0.07in}
\noindent\textbf{Parallelizing an entire network:} 
Consider a simple neural with two FC layers parallelized using
Algorithm~\ref{alg:3d-tensor}. The output $O$ of the first layer would be the
input to the other. However, notice in
Figure~\ref{fig:schematic-agarwal-data-dist} how $O$ is divided across the 3D
tensor parallel grid differently than the input $I$. So to ensure that the
second layer can work with $O$ we would need to transpose its weight matrix --
essentially dividing its rows across the $X$-axis and columns across the
$Y$-axis. This transpose needs to be done once at the beginning of training.
So, to parallelize a full neural network, we simply `transpose' the weights of
every alternate layer by swapping the roles of the $X$- and $Y$- tensor
parallel groups.

\section{Performance Optimizations in \method}
\label{sec:opt}
In this section, we present communication optimizations  that target the
reduction of communication time for a given training workload, consisting of
the neural network, its hyper parameters, and the number of GPUs, along with a
predefined set of values for the configurable performance parameters of our
algorithm. As a running example, we consider a 20B parameter GPT style
transformer~\cite{gpt-3} with a batch size of 32k tokens (sequence length of 2k
tokens) on 16 GPUs or four nodes of the Perlmutter supercomputer. We use a
configuration of $G_{x}=2$, $G_{y}=2$, $G_{z}=4$, and $G_{data}=1$

\subsection{Overlapping All-Reduces with Computation}
\label{sec:opt-row-col-ar}

\begin{figure}[h]
    \centering
      \includegraphics[width=\columnwidth]{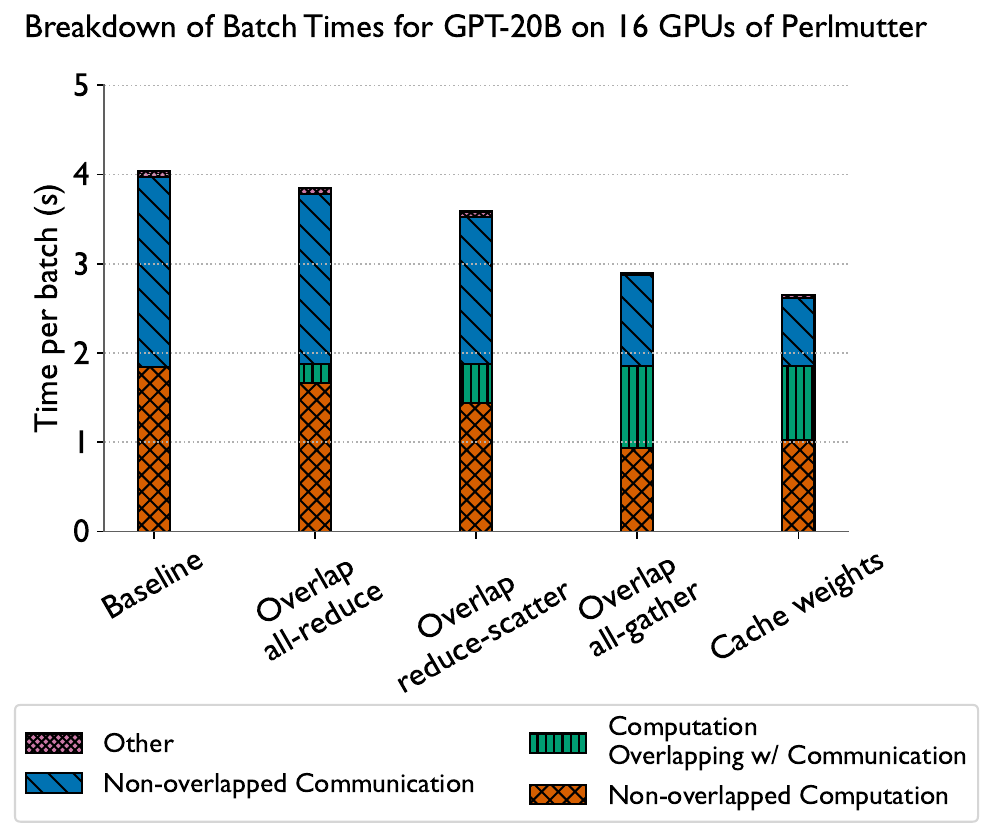}
      \caption{Studying the effect of the proposed communication optimizations on the training times 
      of a GPT 20B model on 16 GPUs of Perlmutter. We use Pipit \cite{bhatele:2023pipit} for 
      creating these breakdowns from trace data collected using the PyTorch Profiler~\cite{pytorch-profiler}.}
\label{fig:opt}
\end{figure}

This optimization is concerned with overlapping the all-reduce communication in
the backward pass of a layer across the $X$-tensor parallel group (Line 11 of
Algorithm~\ref{alg:3d-tensor}) with computation. Note that for layers with
`transposed' weight matrices discussed in the previous section, this
communication would happen across the $Y$-tensor parallel groups. Our strategy
to achieve overlap is to issue the all reduce in line 11 asynchronously and
overlap it with the computation of the weight gradients happening in line 12.
Once this computation has finished, we wait on the asynchronous all reduce to
finish. From Figure~\ref{fig:opt}, we can see that adding this optimization
increases the proportion of communication overlapped with computation,
improving batch times by around 5\%.

\begin{figure}[h]
  \centering
    \includegraphics[width=\columnwidth]{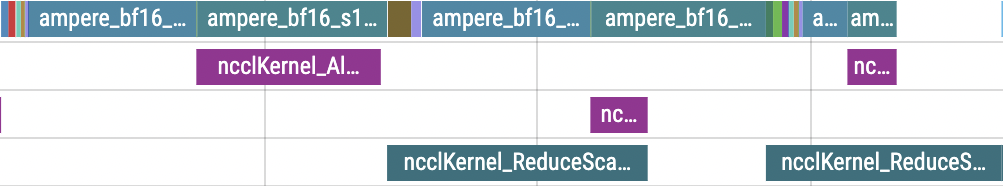} \\
    \vspace{1em}
    \hrule
    \vspace{1em}
    \includegraphics[width=\columnwidth]{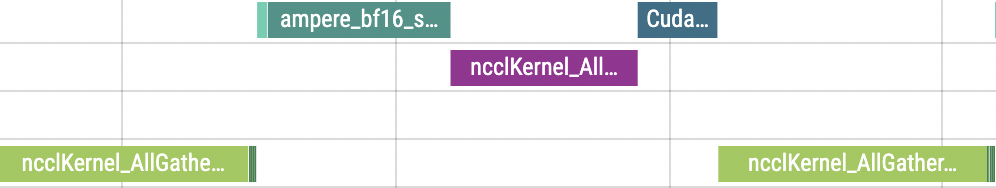} \\
    \vspace{1em}
    \hrule
    \vspace{1em}
    \includegraphics[width=\columnwidth]{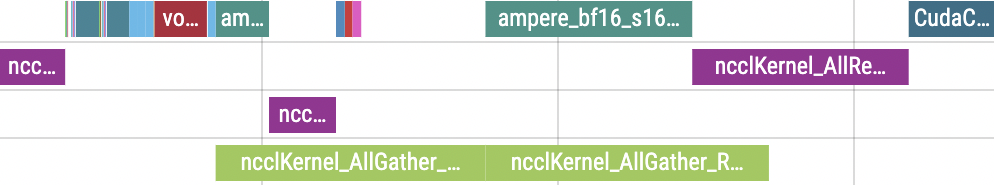}
    \vspace{0.005em}
    \hrule
    \vspace{0.5em}
    \includegraphics[width=\columnwidth]{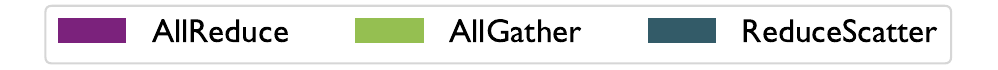}
    \caption{PyTorch Profiler traces demonstrating 
    i.~(top) overlap of reduce scatters with backward pass compute as discussed in Section~\ref{sec:opt-depth-rs},
    ii.~(middle) all-gathers without any overlap in the forward pass, and 
    iii.~(bottom) all-gathers after introducing the overlap optimization in Section~\ref{sec:opt-depth-ag}. 
    The first row in every trace corresponds to the compute stream and the others are communication streams.}
\label{fig:opt-overlap-temp}
\end{figure}

\subsection{Overlapping Reduce-Scatters with Computation}
\label{sec:opt-depth-rs}

Next we look at optimizing the reduce scatters in the backward pass (Line 12 of
algorithm~\ref{alg:3d-tensor}). The outputs of this reduce scatter are the
gradients of the loss w.r.t. the weights of the layer. Note that these aren't
required until we have finished the backward pass of the entire network and are
ready to do the all-reduces pertaining to data parallelism.  Taking advantage
of this we (i) issue these reduce scatters asynchronously and (ii) only wait on
them to complete once all layers have finished their backward pass. This allows
us to overlap the reduce scatter of one layer with the backward pass compute of
its predecessors. From Figure~\ref{fig:opt}, we can see that adding this
optimization further increases the communication-computation overlap and
improves batch times by 8\%. Figure~\ref{fig:opt-overlap-temp} (top) shows a
PyTorch Profiler trace demonstrating that our optimization indeed leads to
overlap of reduce-scatters with computation.

\subsection{Overlapping All-Gathers with Computation}
\label{sec:opt-depth-ag}

Our next optimization aims to overlap the all-gather operations in the forward
pass (Line 2 of Algorithm~\ref{alg:3d-tensor}) with computation. It's important
to note that this all-gather operation doesn't depend on any intermediate
outputs of the forward pass.  Leveraging this independent nature of the
all-gather, we propose to preemptively enqueue the all-gather operation for the
next layer while the computation for the current layer is ongoing. At the
outset of training, we generate a topological sort of the neural network
computation graph to determine the sequence for performing the all-gathers.
Subsequently, we execute them preemptively as outlined earlier. In
Figure~\ref{fig:opt}, we observe that overlapping the all-gathers in this
fashion leads to even more communication-computation overlap and significant
improvement to batch times of nearly 19\%! Figure~\ref{fig:opt-overlap-temp}
shows a PyTorch Profiler trace without (middle) and with (bottom) this
optimization, evidently showing the overlap of all-gathers with forward pass
compute when the optimization is applied.

\subsection{Caching Outputs of All-Gathers}
\label{sec:opt-depth-stash}

Finally, we exploit the fact that most large scale training runs involve
activation checkpointing~\cite{chen2016training}, which basically is a method
to significantly reduce activation memory usage albeit at an effective cost of
an extra forward pass through the network. Since the parameters across the two
forward passes are not changing, the all-gathers in line 2 produces the same
output in the two forward passes. To eliminate the second all-gather we propose
to cache the outputs of the first all-gather in the GPU memory, and reuse them
during the second forward pass. For our running example of the 20B model, we
cache the all-gather outputs of 28 out of the 32 transformer encoder layers and
observe an improvement of nearly 9\%. \emph{Overall, the four optimizations
proposed in this section improve the batch times by a significant 34\%}!


\section{A Performance Model of Collective Communication in Parallel Training} \label{sec:comm-model}
In this section, we address the following question: \emph{how can we configure
the four dimensions (i.e. $G_{x}, G_{y}, G_{z}, G_{data}$) of our 4D algorithm
to minimize total communication time for a given training task}?  To streamline
this process for the end-user, we develop a performance model that predicts
communication time based on the neural network architecture, training
hyperparameters, the four configurable performance parameters, and network
bandwidths. This model returns a small set of near-optimal configurations to
the user, significantly reducing the time needed to discover such
configurations when compared to an exhaustive sweep of the 4D search space. 

\subsection{Placement-agnostic Performance Model}

As detailed in Algorithm~\ref{alg:3d-tensor}, our proposed approach relies on
several collective communication operations, namely all-reduces,
reduce-scatters, and all-gathers, thus we focus on predicting their times. Now,
let us begin by discussing the assumptions we make in our communication model.

\begin{itemize}
    \item \emph{Assumption-1:} The underlying communication libraries use the
bandwidth-optimal ring algorithm~\cite{thakurimproving2003,
rabenseifneroptimization2004} for all-reduce, reduce-scatter, and all-gather
collectives. Note that the ring algorithm is a readily available option within
the NCCL (NVIDIA) and RCCL (AMD) libraries.
    \item \emph{Assumption-2:} For inter-node collectives, the ring is formed
such that the number of ring links crossing node boundaries is minimized.
    \item \emph{Assumption-3:} The message sizes are large enough such that the
message startup overheads can be ignored. 
    \item \emph{Assumption-4:} We are only modeling the communication times and
ignore the effects of any computation taking place on the GPUs.
    \item \emph{Assumption-5:} We assume the same peer-to-peer bidirectional
bandwidth, $\beta_{\mathrm{inter}}$, between every pair of nodes. Similarly, we
assume $\beta_{\mathrm{intra}}$ to be the peer-to-peer bandwidth between two
GPUs within a node.
\end{itemize}

Let us assume that $G$ is the number of GPUs, $\beta$ is the peer-to-peer
network bandwidth, and $m$ is the size of the input buffers being sent from
each GPU. We can write the message transmission times for all-gather
($t_{\mathrm{AG}}$ ), reduce-scatter ($t_{\mathrm{RS}}$), and all-reduce
($t_{\mathrm{AR}}$) as follows: 
\begin{align}
    t_{\mathrm{AG}} &= \frac{1}{\beta}\times (G-1) \times m                           \label{eqn:ring-ag} \\
    t_{\mathrm{RS}} &= \frac{1}{\beta}\times \left( \frac{G-1}{G} \right) \times m    \label{eqn:ring-rs} \\
    t_{\mathrm{AR}} &= \frac{2}{\beta}\times \left( \frac{G-1}{G} \right) \times m    \label{eqn:ring-ar} 
\end{align}

Note that these equations are adapted from the discussion on ring algorithms in
Thakur et al.~\cite{thakurimproving2003} and
Rabenseifner~\cite{rabenseifneroptimization2004}, wherein we ignore the latency
and computation costs in line with our assumptions.

Next, we will use these equations to estimate the time spent in communication
by our 4D algorithm.  Let $t_{\mathrm{AG},z}$ denote the time spent in the
all-gather across the $Z$-tensor parallel groups (line 2 of
Algorithm~\ref{alg:3d-tensor}). Similarly, we use $t_{\mathrm{RS},z}$,
$t_{\mathrm{AR},y}$ and $t_{\mathrm{AR}, x}$ to refer to the time spent in the
collectives in lines 12, 4, and 11 respectively. Similarly, we use
$t_{\mathrm{AR}, \mathrm{data}}$ for the time spent in the data parallel
all-reduce.  Substituting the values of $m$ and $G$ for these operations in
Equations~\ref{eqn:ring-ag} to~\ref{eqn:ring-ar} yields: 

\begin{align}
    t_{\mathrm{AG},z} &= \frac{1}{\beta}\times (G_{z}-1) \times \frac{k \times n}{G_{x} \times G_{y} \times G_{z}}  
    \label{eqn:layer-ag} \\ \nonumber \\
    t_{\mathrm{RS},z} &= \frac{1}{\beta}\times \left( \frac{G_{z}-1}{\mathrm{G_{z}}} \right) \times \frac{k \times n}{G_{x} \times G_{y}} 
    \label{eqn:layer-rs} \\ \nonumber \\
    t_{\mathrm{AR},y} &= \frac{2}{\beta} \times \left( \frac{G_{y}-1}{G_{y}} \right) \times \frac{m \times n}{G_{z} \times \mathrm{G}_{x}}
    \label{eqn:layer-ar-1} \\ \nonumber \\
    t_{\mathrm{AR}, x} &= \frac{2}{\beta} \times \left( \frac{G_{x}-1}{G_{x}} \right)
    \times \frac{m \times k}{G_{z} \times G_{y}}
    \label{eqn:layer-ar-2} \\ \nonumber \\
    t_{\mathrm{AR}, \mathrm{data}} &= \frac{2}{\beta} \times \left( \frac{G_{\mathrm{data}}-1}{G_{\mathrm{data}}} \right) 
    \times \frac{k \times n}{G_{x} \times G_{y} \times G_{z}}
    \label{eqn:layer-ar-3}  
\end{align}

The total communication time for a single layer, ${t_{\mathrm{comm}}}$ is
simply the sum of Equations~\ref{eqn:layer-ag} through~\ref{eqn:layer-ar-3}: 
\begin{align}
    t_{\mathrm{comm}} = t_{\mathrm{AG},z} + t_{\mathrm{RS},z} +  t_{\mathrm{AR},y} + t_{\mathrm{AR},x} 
    + t_{\mathrm{AR}, \mathrm{data}} \label{eqn:layer} 
\end{align}
For layers with `transposed' weight matrices as discussed at the end of
Section~\ref{sec:design}, we need to swap the values of $G_x$ and $G_y$, and
$\beta_{x}$ and $\beta_{y}$. And finally to model the communication time for
the entire model, we apply Equation~\ref{eqn:layer} to all of its layers, and
take a sum of the times.

\subsection{Placement-aware Performance Model}

In the previous section, we made a simplifying assumption that all collectives
in our hybrid parallel method have the same peer-to-peer bandwidth, denoted by
${\beta}$. However, the actual bandwidth available for communication between
peers depends on how we map the process groups of our 4D parallel algorithm
onto the underlying topology. For example, process groups that are contained
entirely within a node can experience higher bandwidths than those containing
GPUs on different nodes, In this section, we attempt to model the specific
bandwidths in Equations~\ref{eqn:layer-ag} through~\ref{eqn:layer-ar-3}, given
$\beta_{inter}$ and $\beta_{intra}$. $\beta_{inter}$ is the peer-to-peer
bandwidth between two GPUs on different nodes (see Assumption 5 at the
beginning of this section); $\beta_{intra}$ is the bandwidth within-node.

To model the process group bandwidths, we begin by assuming a hierarchical
organization of process groups: $X$-tensor parallelism (innermost), followed by
$Y$-tensor parallelism, $Z$-tensor parallelism, and data parallelism
(outermost). As a concrete example, if we have eight GPUs, and set
$G_{x}=G_{y}=G_{z}=G_{\mathrm{data}}=2$, then the $X$-tensor parallel groups
comprise of GPU pairs (0,1), (2,3), (4,5), and (6,7). Similarly, the $Y$-tensor
parallel groups would comprise of GPU pairs (0,2), (1,3), (4,6), and (5,7), and
so on.

Now let $\vec{G} = (G_{x}, G_{y}, G_{z}, G_{\mathrm{data}})$ be the tuple of
our configurable performance parameters, arranged in order of the assumed
hierarchy. Let $\vec{\beta} = (\beta_{x}, \beta_{y}, \beta_{z},
\beta_{\mathrm{data}})$ be the effective peer-to-peer bandwidths for
collectives issued within these process groups. We use $\vec{\beta}_{i}$ and
$\vec{G}_{i}$ to represent the $i^{\mathit{th}}$ elements of these tuples ($0
\leq i \leq 3$). Also, let $G_{\mathrm{node}}$ refer to the number of GPUs per
node. Now let us attempt to model each $\beta_{i}$ i.e. the bandwidth available
to the GPUs in the process groups at the $i^{\mathit{th}}$ level of the
hierarchy.

\subsubsection{Case 1: GPUs in the process group lie within a node}
In terms of our notation, this is the scenario when $\prod_{j=0}^{i}G_{j} \leq
G_{\mathrm{node}}$. In this case we simply use $\beta_{\mathrm{intra}}$ as the
value for $\beta_{i}$.

\subsubsection{Case 2: GPUs in the process group are on different nodes}
In terms of our notation, this is the scenario when $\prod_{j=0}^{i}G_{j} >
G_{\mathrm{node}}$. Let us understand this case with two illustrative examples.

\begin{figure}[h]
    \centering
       \includegraphics[width=\columnwidth]{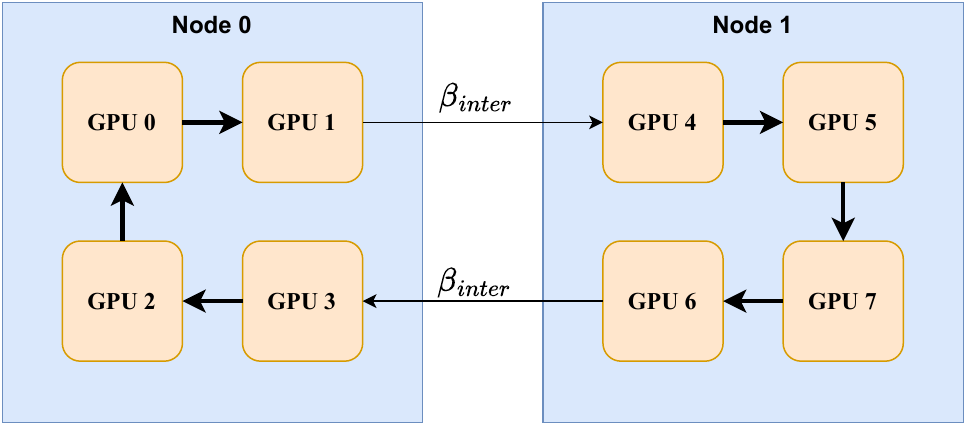}
       \caption{Collective communication operations (all-reduce/reduce-scatter/all-gather) using the ring algorithm, spanning eight GPUs on two nodes.
       \label{fig:all-reduce-one}} 
\end{figure}

In Figure~\ref{fig:all-reduce-one}, we demonstrate a scenario with a single
process group spanning eight GPUs on two nodes, with four GPUs on each node. In
this case, the ring links crossing node boundaries (i.e. the link between GPUs
1 and 4, and the link between GPUs 6 and 3) will be the communication
bottleneck. Since we assumed $\beta_{\mathrm{inter}}$ to be the bidirectional
bandwidth between node pairs, we can set $\beta_{i}=\beta_{\mathrm{inter}}$.

\begin{figure}[h]
    \centering
       \includegraphics[width=\columnwidth]{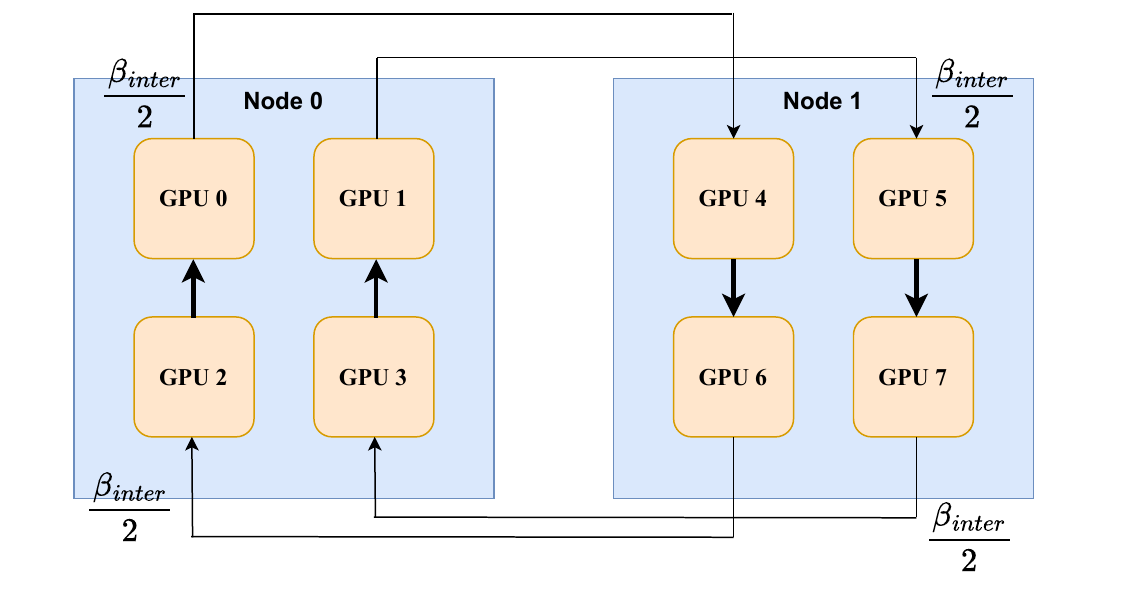}
       \caption{ Two simultaneous communication operations (all-reduce/reduce-scatter/all-gather) using the ring algorithm, each spanning 
       four GPUs. \label{fig:all-reduce-many}} 
\end{figure}

Another possible scenario is when there are multiple simultaneous collectives
taking place between two nodes. For example, consider
Figure~\ref{fig:all-reduce-many}, wherein GPUs $(0, 4, 6, 2)$ and GPUs $(1, 5,
7, 3)$ are executing two collectives with the ring algorithm simultaneously. In
this case, the available inter-node bandwidth will be shared between these two
collectives and $\beta_{i}=\frac{\beta_{\mathrm{inter}}}{2}$. 

The first scenario occurs in the case when the process groups preceding the
$i^{\mathit{th}}$ process group in the hierarchy are of size one, i.e.
$G_{j}=1$  $\forall j < i$. Whereas the second scenario occurs in the case when
at least one of these preceding process groups is of a size $>1$. In that case,
we get multiple ring links crossing node boundaries and the bandwidth gets
divided between the rings. However, note that the maximum reduction in the
bandwidth is bounded by the total number of GPUs on each node, as there can't
be more inter-node ring links than GPUs on a node. Equation~\ref{eqn:bandwidth}
models all the scenarios to obtain the observed bandwidth:
\begin{equation}
\vec{\beta}_{i} = \
         \dfrac{\beta_{\text{inter}}}{\min \left( G_{\text{node}}, \prod_{j=0}^{i-1}G_{j} \right)} \label{eqn:bandwidth}
\end{equation}
When we use this bandwidth term in our model, we refer to it as the
placement-aware performance model.

\subsection{Validating the Performance Models}

\begin{figure}[h]
    \centering
        \includegraphics[width=0.49\columnwidth]{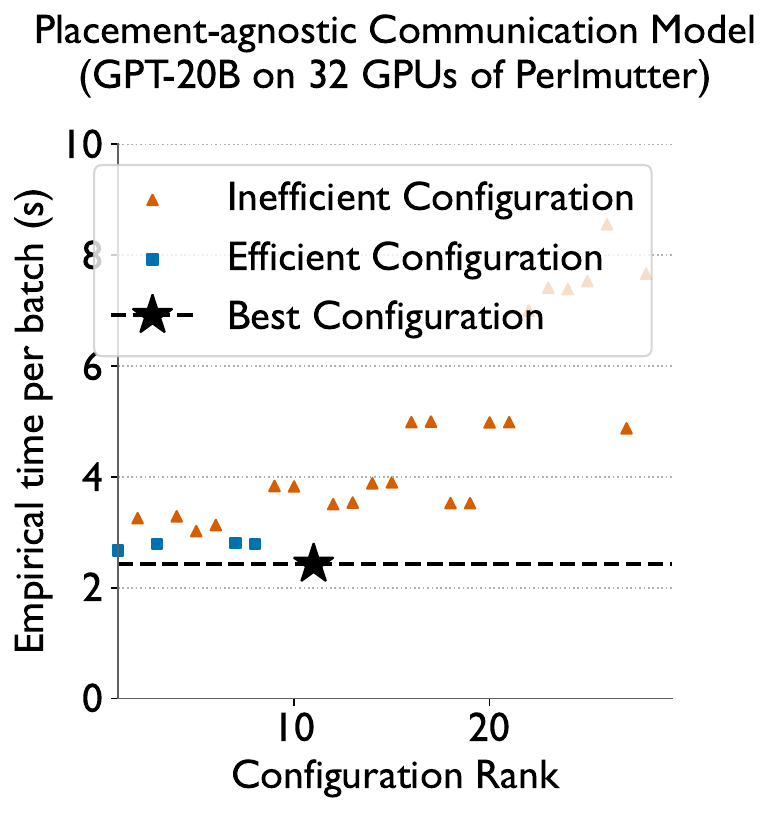}
        \includegraphics[width=0.49\columnwidth]{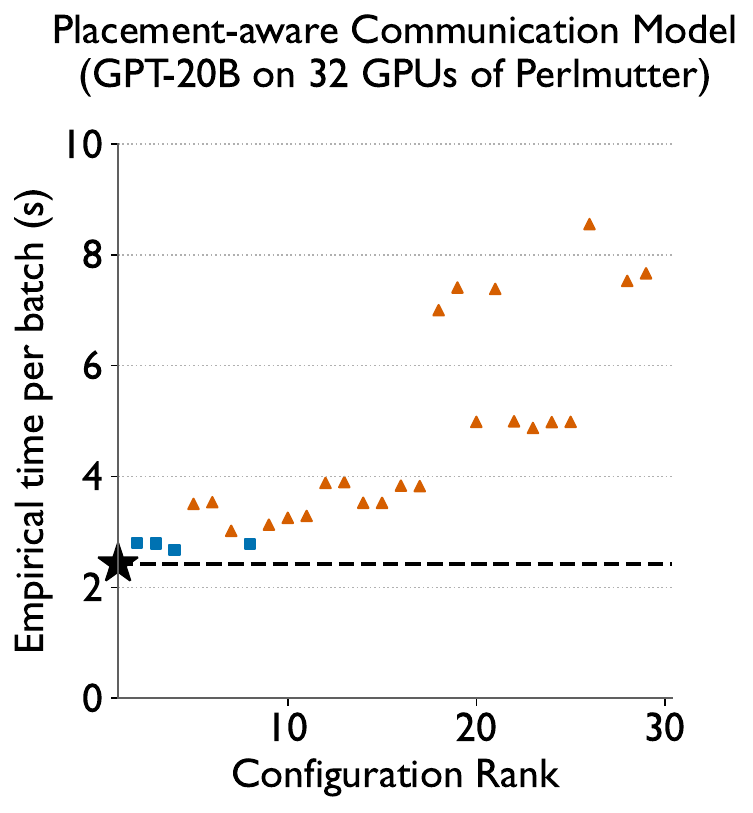}
        \caption{Comparison of empirical batch times across different configurations, ranked by both 
        placement-agnostic (left) and placement-aware communication models (right). \label{fig:rank-comm-model}}
\end{figure}

To compare the two performance models, we collect the batch times for all
possible parallel configurations of \method when training GPT-20B on 32 GPUs on
Perlmutter. We classify the top five configurations with respect to the batch
times as `efficient' and the rest as `inefficient'. We then rank these
configurations using both the placement-agnostic (setting all bandwidths to a
constant value) and placement-aware performance models. In
Figure~\ref{fig:rank-comm-model}, we show the empirical batch times with
respect to the configurations ranked by the placement-agnostic version (left)
and the placement-aware version (right) of our model.  Notice how four out of
the top five configurations identified by the placement-aware model are
`efficient', while the placement-agnostic model only recognizes two.
Additionally, the optimal configuration ranks first in the placement-aware
model but twelfth in the placement-agnostic model. This experiment demonstrates
the utility of modeling the network bandwidths in our communication model, as
well as the effectiveness of the placement-aware model in identifying efficient
configurations.

\section{Experimental Setup}
\label{sec:setup}
This section provides a detailed account of our empirical evaluation of the
proposed 4D hybrid parallel algorithm which we have integrated in \method.  Our
experiments were conducted on two supercomputers, Perlmutter and Frontier. On
Perlmutter, each node is equipped with four NVIDIA A100 GPUs, each with a DRAM
capacity of 40GB. Whereas on Frontier, each node has four AMD Instinct MI250X
GPUs each with a DRAM capacity of 128GB. Each MI250X GPU is partitioned into
two Graphic Compute Dies (GCDs) and each GCD appears as a separate device to
any deep learning framework. Nodes on both systems have four HPE Slingshot 11
NICs, with each NIC capable of link speeds of 200 Gb/s. 

\subsection{Description of Neural Networks and Hyper-parameters} \label{sec:setup-desc}
  
We evaluate the effectiveness of our proposed framework by conducting
experiments on two well-known neural network architectures:
U-Net~\cite{unet-arch} and GPT~\cite{gpt-3}. U-Nets are fully convolutional
neural networks that have diverse applications in various fields such as
text-to-image systems (e.g., Dall-E-2~\cite{dall-e-2} and
Stable-Diffusion~\cite{stable-diffusion}), image
segmentation~\cite{image-segmentation-survey}, and object
detection~\cite{object-detection-survey}. The GPT architecture is a popular
transformer architecture~\cite{transformer} that has been used to develop
several large language models~\cite{megatron-turing-nlg-530b, bloom176b, gpt-3,
gpt-2}.  Tables~\ref{tab:setup-perf-gpt} and ~\ref{tab:setup-perf-unet} detail
the model architectures and their corresponding hyperparameters. Due to the
extremely large activation memory requirements of training GPT models, we turn
on activation checkpointing~\cite{chen2016training}.  Additionally, we employ
mixed precision (bf16/fp32) for all our training runs. 

\begin{table}[h]
  \centering
  \caption{\label{tab:setup-perf-gpt} Architectural details of the GPT-style transformers~\cite{gpt-3} that we use in 
  this work.
  }
  \begin{tabular}{lrrrr}
  \toprule
  Model      &\# Layers  & Hidden-Size &\# Heads  \\ \midrule
  GPT-5B     &   24      & 4096        & 32   \\
  GPT-10B    &   32      & 5120        & 40   \\
  GPT-20B    &   32      & 7168        & 56   \\ 
  GPT-40B    &   38      & 9216        & 72    \\
  GPT-80B    &   42      & 12288       & 96    \\
  \bottomrule
  \end{tabular}
\end{table}

\begin{table}[h]
  \centering
  \caption{\label{tab:setup-perf-unet} List of U-Net models~\cite{unet-arch} that we employed in our weak scaling 
  experiments on Perlmutter. Consistent with Nichol et al.~\cite{improved-diffusion}, our models consist of four levels, 
  with each level comprising three residual blocks. For training, we set the batch size to 2048 and the image 
  resolution to $32\times32$.
  }
  \begin{tabular}{lrr}
  \toprule
   Model & Channels & \# GPUs  \\ \midrule
  U-Net 250M          & 256       & 64     \\
  U-Net 500M          & 416       & 128     \\
  U-Net 1B            & 512       & 256     \\ 
  U-Net 2B            & 768       & 512   \\
  U-Net 4B            & 1024      & 1024   \\
  \bottomrule
  \end{tabular}
\end{table}

To create parallel implementation of these architectures with \method, we
started with established sequential implementation. For U-Nets, we parallelized
Nichol et al.'s sequential implementation~\cite{improved-diffusion} using our
4D approach. For transformers on Perlmutter, we leveraged the optimized
sequential code from the Megatron-LM codebase~\cite{megatronlm-2}. However,
when evaluating on Frontier, we encountered training instability with
Megatron-LM and switched to LitGPT~\cite{litgpt-2023} as the foundation. We
then successfully integrated our 4D algorithm as its backend.

To validate the correctness of our implementation, we train a small 50M
parameter U-Net on the CIFAR-10 dataset~\cite{cifar-10} for 12,000 iterations
as well as a 125M parameter GPT on the BookCorpus dataset~\cite{book-corpus}
for up to 14,000 iterations and present the training losses for both of them.
We then conduct weak scaling experiments with the U-Net, starting from a 250M
parameter model on 64 GPUs (64 GCDs on Frontier), and scaling up to 4B
parameters on 1024 GPUs (1024 GCDs on Frontier).  We conduct a similar weak
scaling experiment with the GPT-3 models, ranging from GPT-5B to GPT-80B on
64-1024 GPUs on Perlmutter (64-1024 GCDs on Frontier).  Note that we had to
make slight adjustments to layers and hidden-sizes of the GPT models listed in
Table~\ref{tab:setup-perf-gpt} for Megatron-LM because it requires the number
of layers to be divisible by the pipeline parallelism dimension.  Additionally,
we conduct a strong scaling experiment for GPT-80B ranging from 64-1024
GPUs/GCDs on each supercomputer.

\subsection{Choice of Frameworks for Comparison}
\label{sec:setup-framework}

We compare the performance of our proposed hybrid parallel framework with three
state-of-the-art baseline frameworks: Megatron-LM~\cite{megatronlm,
megatronlm-2}, DeepSpeed-3D~\cite{deepspeed-extreme-3d};
ZeRO-3~\cite{sc2020zero}. Megatron-LM combines tensor, pipeline and data
parallelism to efficiently train large multi-billion parameter GPT-style
transformers at scale.  Like Megatron-LM, DeepSpeed-3D combines data
parallelism, pipeline parallelism, and tensor parallelism. ZeRO-3 is stage 3 of
the ZeRO optimizer which partitions the optimizer states, gradients, and
parameters across GPUs. Note that we do not run Megatron-LM on Frontier due to
the training instabilities mentioned in the previous subsection.  Note that
since Megatron-LM and DeepSpeed-3D do not provide parallel implementations of
UNets, we only run ZeRO-3 as our baseline for the weak scaling experiments on
UNets.

\subsection{Evaluation Metrics}

For our weak and strong scaling experiments we report the average time per
iteration. We do so by running each framework for ten batches and reporting the
average of the last five. For our GPT runs, we also calculate half precision
FLOP/s using Narayanan et al.~'s~\cite{megatronlm-2} analytical formulation for
the number of floating point operations in a transformer. We then compare this
number with the theoretical peaks on each machine (312 TFLOP/s per GPU on
Perlmutter, and 192 TFLOP/s per GCD on Frontier) and report the achieved
percentage of peak.


\section{Scaling Results}
\label{sec:results}
In this section, we describe the results of the empirical experiments outlined
in Section~\ref{sec:setup}.

\subsection{Validating Our Implementation}

\begin{figure}[h]
  \centering
    \includegraphics[width=0.49\columnwidth]{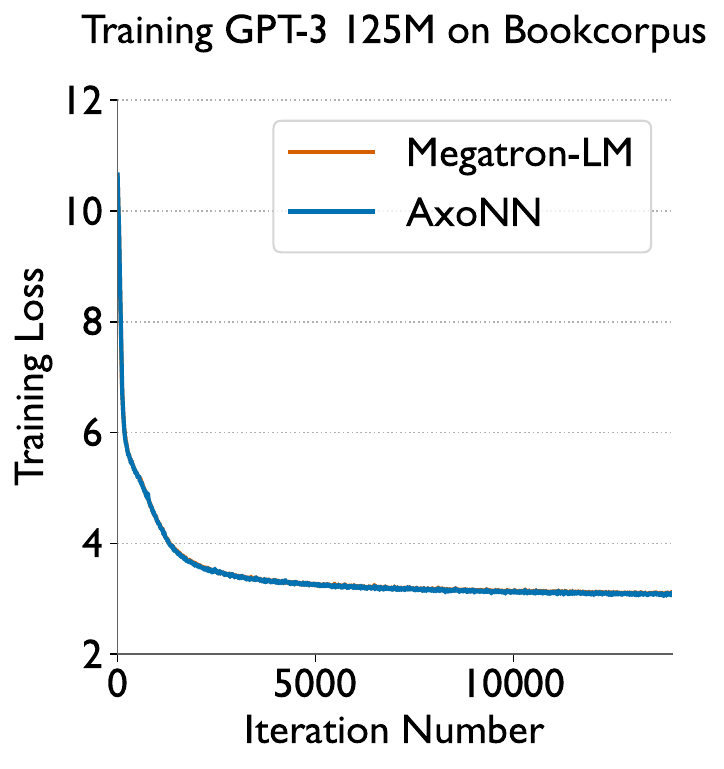}
    \includegraphics[width=0.49\columnwidth]{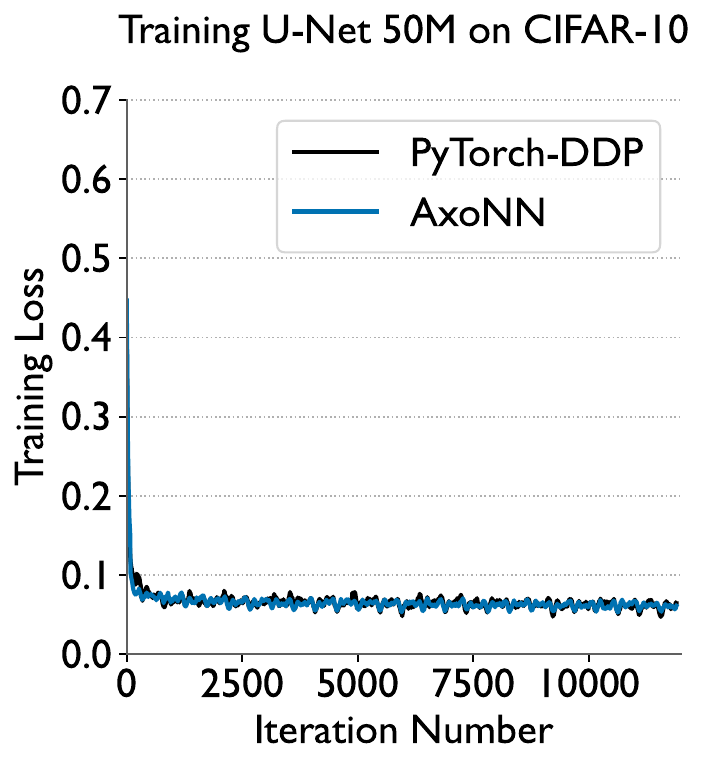}
    \caption{ Validating the correctness of our implementation by training GPT-125M and UNet-50M to completion.
    Details about the training hyperparameters can be found in Section~\ref{sec:setup-desc}.}
\label{fig:stat-eff}
\end{figure}

To establish the correctness of our implementation, we present the loss curves
for a 125M parameter GPT and a 50M parameter UNet model trained on 16 GPUs
using \method in the left and right sides of Figure~\ref{fig:stat-eff}
respectively.  For both the experiments, we set $G_x=G_y=G_z=G_{data}=2$ so
that each dimension in our algorithm is active. We also switch on all of the
communication optimizations discussed in Section~\ref{sec:opt}. 

For the GPT experiment, we compare with Megatron-LM in a pure data parallel
configuration on sixteen GPUs. We observe that \method successfully trains the
model to convergence and produces near identical loss curves with Megatron-LM,
thus validating our implementation. For the UNet experiment, we compare with
PyTorch-DDP~\cite{pytorchdist-vldb}, which is an implementation of data
parallelism native to PyTorch. Again, we observe that \method successfully
trains the model to convergence and produces near identical loss curves. 

\begin{figure*}[t]
  \centering
    \includegraphics[width=\columnwidth]{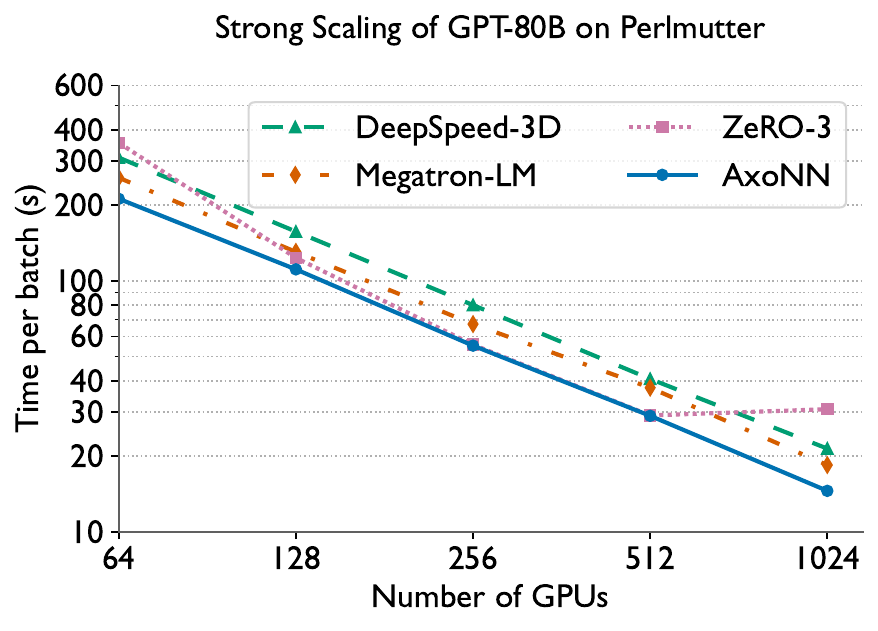}
    \includegraphics[width=\columnwidth]{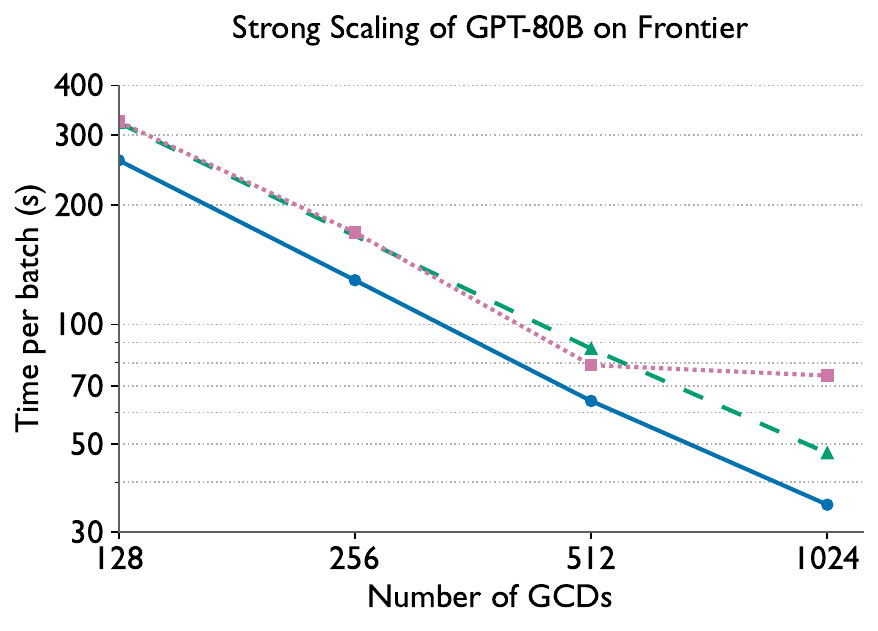}
    \caption{Time per batch for strong scaling of GPT-80B on Perlmutter (left) and Frontier (right). 
    We use a batch size of 4M tokens and a sequence length of 2048.~\label{fig:strong-scale-gpt}}
\end{figure*}

\subsection{Strong Scaling}

Next, we demonstrate the results of our strong scaling experiments on the GPT
80B architecture in Figure~\ref{fig:strong-scale-gpt} on Perlmutter (left) and
Frontier (right). On Perlmutter, we observe that \method, Megatron-LM, and
DeepSpeed-3D scale linearly upto 1024 GPUs. ZeRO-3 scales extremely well upto
512 GPUs matching \method's iteration times, but degrades significantly at 1024
GPUs. On 1024 GPUs, \method outperforms Megatron-LM by a significant 25\%, and
DeepSpeed-3D by 32\%! This result is particularly noteworthy considering
Megatron-LM and DeepSpeed-3D, both of which combine tensor, pipeline, and data
parallelism, are two of the leading approaches in parallel deep learning. Both
have been instrumental in training numerous large language models (LLMs) in
real-world applications~\cite{megatron-turing-nlg-530b, bloom176b,
parmar2024nemotron4, black-etal-2022-gpt}. Next, let us look at the strong
scaling results on Frontier (right of Figure~\ref{fig:strong-scale-gpt}). Once
again we observe that ZeRO-3 does not scale beyond 512 GCDs, whereas
DeepSpeed-3D and \method demonstrate near linear scaling upto 1024 GCDs. On
1024 GCDs, \method is faster than DeepSpeed-3D by 26\% and ZeRO-3 by nearly
52\%!  

\begin{figure}[h]
  \centering
    \includegraphics[width=\columnwidth]{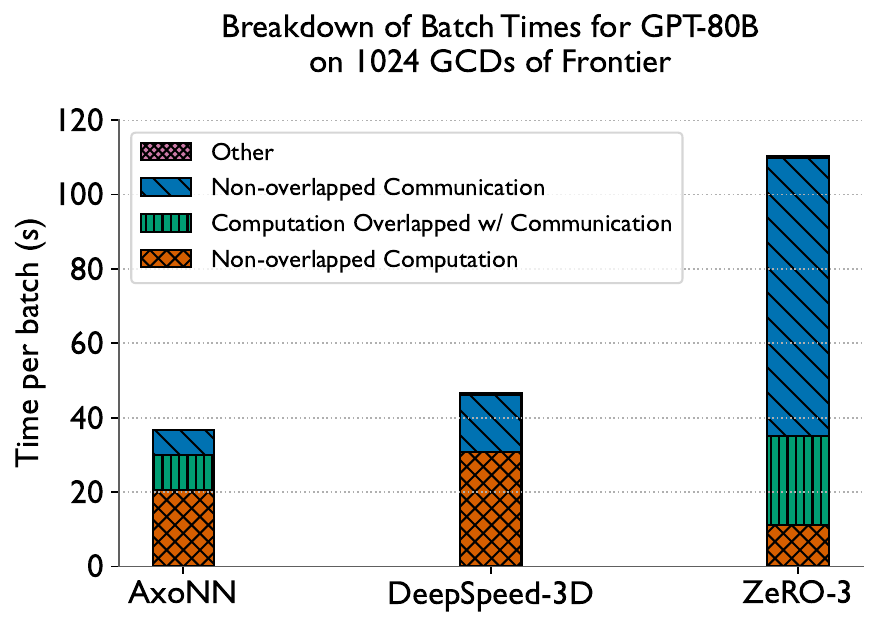}
    \caption{Comparison of breakdown of batch times for different frameworks on a 80B GPT model run on 1024 GCDs of Frontier. 
    \label{fig:framework-breakdown}} 
\end{figure}

As established in Sections~\ref{sec:opt} and~\ref{sec:comm-model}, our primary
objective has been to minimize the expensive overheads of communication. To
demonstrate that our performance gains stem directly from this focus,
Figure~\ref{fig:framework-breakdown} presents a detailed breakdown of batch
times for the 80B parameter model running on 1024 GCDs of Frontier. First note
that all three of \method, DeepSpeed-3D, and ZeRO-3 spend nearly the same
amount of time in computation - which is the sum of non-overlapped computation
(red) and computation overlapped with communication (green). However, when it
comes to time spent in non-overlapped communication (blue), we notice
significant differences. \method only spends 6 seconds of the batch time in
non-overlapped communication, which is nearly $2.3\times$ smaller than
DeepSpeed-3D (15.2 seconds), and nearly $11\times$ smaller than ZeRO-3 (74.9
seconds). This clearly demonstrates the effectiveness of our approach in
minimizing communication overheads, leading to the observed performance gains.

\begin{figure*}[t]
  \centering
    \includegraphics[width=\columnwidth]{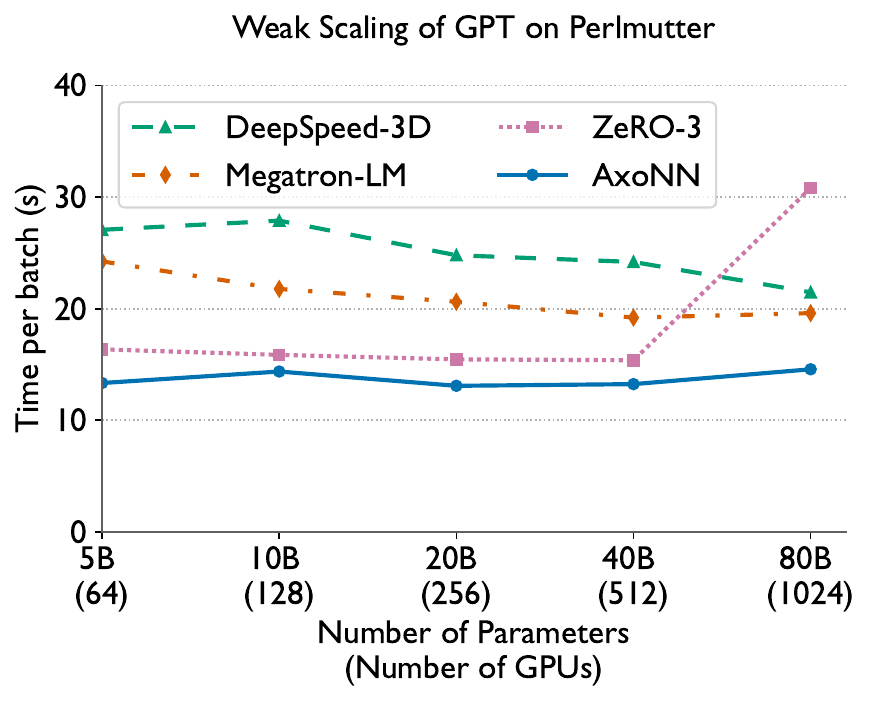}
    \includegraphics[width=\columnwidth]{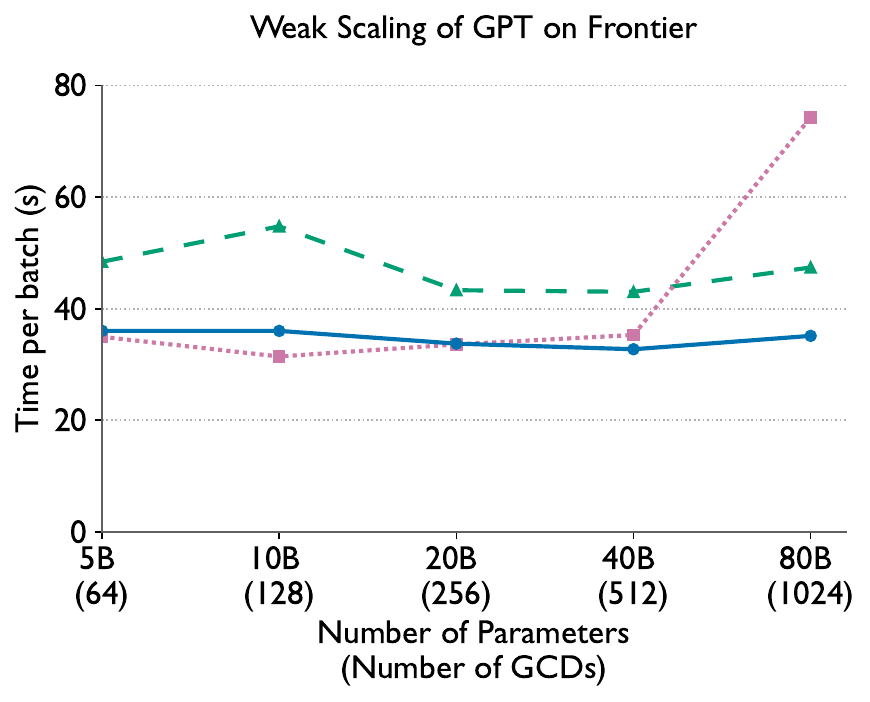}
    \caption{Time per batch for weak scaling of GPT transformers on Perlmutter (left) and Frontier (right). 
    We use a batch size of 4M tokens and a sequence length of 2048.~\label{fig:weak-scale-gpt}} 
\end{figure*}

\subsection{Weak Scaling}

Now let us turn our attention to the weak scaling experiments, starting with
the GPT architecture.  We compare the time per iteration (or batch) for
\method, Megatron-LM, ZeRO-3, and DeepSpeed-3D on Perlmutter in
Figure~\ref{fig:weak-scale-gpt} (left). On Perlmutter, we observe that \method
has the lowest time per iteration for all models and GPU counts. For instance,
\method shows improvements in the range of 25--45\% over Megatron-LM. For GPT
10B, 20B, and 40B, \method performs better than the second best performing
method ZeRO-3, with improvements in the range of 10--18\%. However, similar to
what we observed in our strong scaling experiments, ZeRO-3 does not scale to
1024 GPUs. At this scale, \method demonstrates a 55\% improvement over ZeRO-3.
Figure~\ref{fig:weak-scale-gpt} (right) shows the performance of \method,
ZeRO-3, and DeepSpeed-3D on Frontier. Similar to Perlmutter, \method
demonstrates the lowest time per iteration for all models and GPU counts on
Frontier for GPT 40B, and 80B. For GPT 5B, 10B, and 20B, ZeRO-3 outperforms
\method by a small margin. However, with increasing GPU counts, ZeRO-3 stops
scaling efficiently, while \method continues to scale efficiently. In terms of
scaling, \method is the best performing method for all models and GPU counts on
Perlmutter and Frontier, followed by Megatron-LM on Perlmutter and DeepSpeed on
Frontier. 

\begin{table}[h]
  \centering 
  \caption{Hardware flop/s utilization for weak scaling of GPTs on Perlmutter.
  \label{tab:peak-flops-gpt-weak-perm}}
  \resizebox{\columnwidth}{!}{
  \begin{tabular}{llrrrr}
  \toprule
   \#GPUs  & Model & Megatron-LM    & ZeRO-3      & DeepSpeed-3D & \method \\ \midrule
      64   & GPT-5B &   37\%        & 55\%        & 33\%       & \textbf{67\%}          \\
      128  & GPT-10B &  42\%        & 57\%        & 33\%      & \textbf{63\%}          \\
      256  & GPT-20B &  42\%        & 57\%        & 35\%      & \textbf{67\%}   \\
      512  & GPT-40B &  44\%        & 55\%        & 35\%      & \textbf{64\%}    \\
      1024 & GPT-80B &  42\%        & 27\%        & 38\%      & \textbf{57\%} \\
  \bottomrule
  \end{tabular}
  }
\end{table}

Table~\ref{tab:peak-flops-gpt-weak-perm} lists the hardware flop/s utilization
for the weak scaling of GPTs on Perlmutter. We notice that \method demonstrates
the highest utilization for almost all models and GPU counts, with a
significantly high 57\% of the peak half precision flop/s at 1024 GPUs of
Perlmutter, which is nearly 16\% of the machine! This is much higher than the
next fastest framework - Megatron-LM, which clocks a significantly lower 42\%
of the peak.

\begin{figure}[h]
  \centering
    \includegraphics[width=\columnwidth]{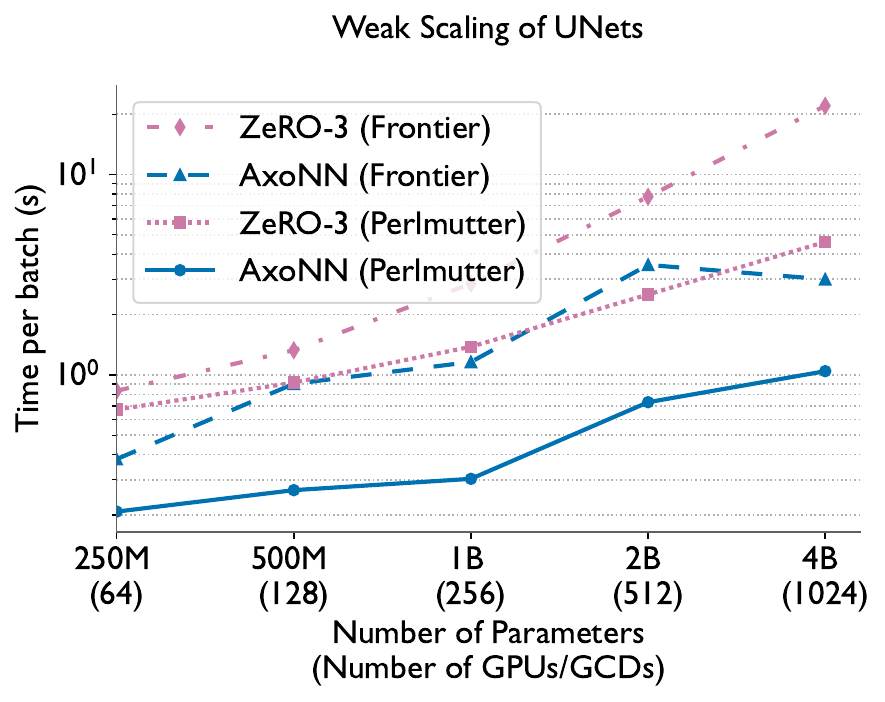}
    \caption{Time per batch for weak scaling of UNets on Perlmutter and Frontier. 
    We use a batch size of 2048 samples. \label{fig:weak-scale-unet}} 
\end{figure}

Now, we turn our attention to the Figure~\ref{fig:weak-scale-unet} which shows
the weak scaling performance of \method and ZeRO-3 for U-Nets on Perlmutter and
Frontier. Again, we observe that \method is significantly faster than ZeRO-3
for all U-Net models and GPU counts. On higher GPU counts of 512 and 1024,
\method is upto 5 times faster than ZeRO-3 on both machines. 

\section{Conclusion}
To overcome communication overheads in parallel deep learning, we introduced a
communication-efficient four-dimensional (4D) hybrid parallel algorithm which
leverages a  variation of Agarwal et al.'s 3D parallel matrix multiplication
algorithm~\cite{agarwal-3d}, but goes beyond that by employing a two-pronged
approach for communication efficiency.  Firstly, we proposed communication
optimizations that exploit asynchronous communication. This allows for
significant overlap between communication and computation, maximizing hardware
utilization during training. Secondly, we introduced a communication model that
identifies a small set of communication-optimal configurations for our
approach.  By combining an efficient parallelization approach with these
communication-centric strategies, \method offers a significant step forward in
tackling the communication bottleneck and enabling efficient large-scale
training of neural networks.


\bibliographystyle{IEEEtran}
\bibliography{./bib/cite,./bib/pssg}


\end{document}